\def\year{2016}\relax
\pdfoutput=1
\documentclass[letterpaper]{article}
\usepackage{aaai16}
\usepackage{times}
\usepackage{helvet}
\usepackage{courier}
\usepackage{amsmath}
\usepackage{multirow}
\usepackage{amsfonts}
\usepackage{dsfont}
\usepackage{color}
\usepackage{graphics}
\newtheorem{theo}{Theorem}

\newtheorem{defi}{Definition}
\newtheorem{lemma}{Lemma}

\def\pf{\noindent {\it Proof.}}
\def\qed{\hfill \rule{4pt}{7pt}}
\begin{document}
%
\title{Locally Adaptive Translation for Knowledge Graph Embedding}
\author{Yantao Jia$^1$, Yuanzhuo Wang$^1$, Hailun Lin$^2$, Xiaolong Jin$^1$, Xueqi Cheng$^1$\\
$^1$CAS Key Laboratory of Network Data Science and Technology,\\
Institute of Computing
Technology,
Chinese Academy of Science, Beijing, China\\
$^2$Institute of Information Engineering, \\
Chinese Academy of Science, Beijing, China\\
}
\maketitle
\begin{abstract}
\begin{quote}
Knowledge graph embedding aims to represent entities and relations in a large-scale knowledge graph as elements in a continuous vector space. Existing methods, e.g., TransE and TransH, learn embedding representation by defining  a global margin-based loss function over the data. However, the optimal loss function is determined during experiments
whose parameters are examined among a closed set of candidates. Moreover, embeddings over two knowledge graphs with different entities and relations share the same set of candidate loss functions, ignoring the locality of both graphs.
This leads to the limited performance of embedding related applications. In this paper, we propose a locally adaptive translation method for knowledge graph embedding, called TransA, to find the optimal loss function by adaptively determining its margin over different knowledge graphs.
Experiments
on two benchmark data sets demonstrate the superiority of the proposed method, as compared to the-state-of-the-art ones.
\end{quote}
\end{abstract}


\section{Introduction}

A knowledge graph is actually a graph with entities of different types as nodes and various relations among them as edges. Typical examples include Freebase \cite{bollacker2008freebase}, WordNet \cite{miller1995wordnet}, OpenKN \cite{jia2014openkn}, to name a few. In the past decade, the great success of constructing these large-scale knowledge graphs have advanced many realistic applications, such as, link prediction \cite{liu2014lsdh} and document understanding \cite{wu2012probase}. However, since knowledge graphs usually contain millions of vertices and edges, any inference and computation over them may not be easy. For example, when using Freebase for link prediction, we need to deal with 68 million of vertices and one billion of edges.  In addition, knowledge graphs usually adopt symbolic and logical representation only. This is not enough when handling applications with intensive numerical computation.

Recently, many research work have been conducted to embed a knowledge graph into a continuous vector space, called knowledge graph embedding, to tackle these problems. For example, TransE~\cite{bordes2013translating} represents entities as points and  relations as translation from head entities to tail entities in the vector space. TransH~\cite{wang2014knowledge} models  relations as translation on a hyperplane and represents a relation as two vectors, i.e., the norm vector of the hyperplane and the translation vector on the hyperplane.
These methods finally learn the representations of entities and relations by minimizing a global margin-based loss function. 

On one hand, existing embedding methods over one knowledge graph determine the optimal form of loss function during experiments. A critical problem is that the determination is made over a limited number of candidates. For example, the optimal loss function used in TransE \cite{bordes2013translating} on Freebase is determined with its margin among the 3-element set $\{1,2,10\}$. It is unclear that why the loss function is examined by only testing a closed set of values of its parameters in the literature. Obviously, the margin is nonnegative and the full grid search of finding its optimum is almost impossible.
On the other hand, existing embedding methods over two different knowledge graphs find their individual optimal loss functions over the same set of candidates. For instance, in TransH~\cite{wang2014knowledge}, the loss functions on Freebase and WordNet share the same candidate margins, i.e., $\{0.25,0.5,1,2\}$. Since different knowledge graphs contain different entities and relations, this compatible setting ignores the individual locality of knowledge graphs and seems not convincing in theory.

In this paper, we propose a translation based embedding method, called TransA, to address the above two issues. For different knowledge graphs, TransA adaptively finds the optimal loss function according to the structure of knowledge graphs, and no closed set of candidates is needed in advance. It not only makes the translation based embedding more tractable in practice, but promotes the performance of embedding related applications, such as link prediction and triple classification.
Specifically, the contributions of the paper are two-fold.
\begin{itemize}
\item We experimentally prove that knowledge graphs with different localities may correspond to different optimal loss functions, which differ in the setting of margins. Then we study how margin affects the performance of embedding by deducing a relation between them.

\item We further propose the locally adaptive translation method (TransA) for knowledge graph embedding. It finds the optimal loss function by adaptively determining the margins over different knowledge graphs.
    Finally, experiments on two standard benchmarks validate the effectiveness and efficiency of TransA.

\end{itemize}

\section{Related Work}

Existing knowledge embedding methods aim to represent entities and relations of knowledge graphs as vectors in a continuous vector space, where they usually define a loss function to evaluate the representations. Different methods differ in the definition of loss functions with respect to the triple $(h,r,t)$ in knowledge graph, where $h$ and $t$ denote the head and tail entities, and $r$ represents their relationship.
The loss function
implies some type of transformation on $h$ and $t$. Among translation based methods, \textbf{TransE} \cite{bordes2013translating} assumes $h+r=t$ when $(h,r,t)$ is a golden triple, which indicated that $t$ should be the nearest neighbor of $h+r$. \textbf{Unstructured} method \cite{bordes2012joint,bordes2014semantic} is a naive version of TransE by setting $r=0$.
To deal with relations with different mapping properties, \textbf{TransH} \cite{wang2014knowledge} was established to project entities into a relation-specific hyperplane and the relation
becomes a translating operation on hyperplane.
Another direction of embedding is the energy based method, which assigns low energies to plausible triples of a knowledge graph and employs neural network for learning. For example,
\textbf{Structured Embedding (SE)} \cite{bordes2011learning} defines two relation-specific matrices for head entity and tail entity, and establishes the embedding by a neural network architecture.
\textbf{Single Layer Model (SLM)} is
 a naive baseline of \textbf{NTN} \cite{socher2013reasoning} by concatenating $h$ and $t$ as an input layer to a non-linear hidden layer.
 Other energy based methods include \textbf{Semantic Matching Energy (SME)} \cite{bordes2012joint}, \textbf{Latent Factor Model (LFM)} \cite{jenatton2012latent,sutskever2009modelling} and \textbf{Neural Tensor Network (NTN)} \cite{socher2013reasoning}. Besides, matrix factorization based methods were also presented in recent studies, such as \textbf{RESCAL} \cite{nickel2012factorizing}. However, these methods find the optimal loss function of embedding whose parameters are selected
 during experiments.
 We intend to propose an adaptive translation method to find the optimal loss function in this paper.

\section{Loss Function Analysis}

In this section, we firstly study the loss functions over  different knowledge graphs and find their difference in the setting of margins. Then in order to figure out how margin affects the performance of embedding, we derive a relation  by virtue of the  notion of
stability for learning algorithms \cite{bousquet2002stability}.

\subsection{Margin setting over different knowledge graphs}

As mentioned before, a knowledge graph is composed of heterogeneous entities and relations.
Knowledge graphs with different types of entities and relations are  different and exhibit different \textit{locality} with regard to the types of their elements.
In this sense, existing knowledge embedding methods using a common margin-based loss function cannot satisfactorily represent the locality of knowledge graphs.

To verify this, we construct knowledge graphs with different locality. We simply partition a knowledge graph into different subgraphs in a uniform manner. Each subgraph contains different types of relations and their corresponding entities. Moreover, different subgraphs have the identical number of relations for the sake of balance of the number of entities.
We claim that over different subgraphs, the optimal margin-based loss function may be different in terms of the margin.
To validate this point, we perform the embedding method TransE \cite{bordes2013translating} on the data set FB15K from the knowledge graph, Freebase. And we partition FB15K into five subsets with equal size of relations.
For example, one subset, named Subset1,  contains 13,666 entities and 269 relations. Another subset, named Subset2, has 13603 entities and 269 relations.
We conduct link prediction task over these five subsets and use
mean rank (i.e., mean rank of correct entities)
to evaluate the results shown in Table \ref{compmar}.

\begin{table}[h,t]
\centering
\setlength{\abovecaptionskip}{4pt}
\setlength{\belowcaptionskip}{-11pt}
\begin{tabular}{|c|c|c|c|}
\hline
\multirow{2}{*}{Data sets} & \multirow{2}{*}{Optimal loss function} & \multicolumn{2}{|c|}{Mean Rank} \\
\cline{3-4}
 & & Raw & Filter \\
\hline
\cline{1-4}
Subset1 & $f_r(h,t)+3-f_{r}(h',t')$ & 339&	240\\
Subset2 & $f_r(h,t)+2-f_{r}(h',t')$  &500&	365\\
\hline
FB15K & $f_r(h,t)+1-f_{r}(h',t')$ & 243 &125\\
\hline
\end{tabular}
\caption{Different choices of optimal loss functions and the predictive performances over three data sets Subset1, Subset2 and FB15K, where $f_r(h,t)=\|h+r-t\|^2_2$, $(h,r,t)$ is a triple in knowledge graph, and $(h',r,t')$ is incorrect triple.
 }\label{compmar}
\end{table}

It follows from Table \ref{compmar} that the settings of loss functions are different over the two data sets, Subset1 and Subset2. More precisely, they take different values of margin as 3 and 2, respectively.
Meanwhile, for the whole data set FB15K, it has been shown in \cite{bordes2013translating} that the best performance is achieved when the optimal loss function takes the margin 1.
This suggests that knowledge embedding of different knowledge graphs with a global setting of loss function can not well represent the locality of knowledge graphs, and it is indispensable to propose a locality sensitive loss function with different margins.

\subsection{How margin affects the performance}

So far we have experimentally found that the performance of a knowledge embedding method is relevant to the setting of margins in the margin-based loss function. This raises a theoretical question that how the setting of margin affects the performance when the margin increases.
To answer this question,  we present a relation between the error of performance and the margin.

Before obtaining the relation, let us firstly give some notations.
Denote the embedding method by $\mathcal{A}$.
Since the embedding method is to learn the appropriate representations of entities and relations in the knowledge graph, it can be viewed as a learning algorithm.
Suppose that for the  embedding method and the knowledge graph, the training data set
\[
S=\{\left(h_1,r_1,t_1\right),\ldots,\left(h_i,r_i,t_i\right),\ldots,\left(h_n,r_n,t_n\right)\}
\]
of size $n$ is a set of triplets
in the knowledge graph
and the range of $\left(h_i,r_i,t_i\right)$ is denoted by $\mathcal{Z}$.
 Denote $S^{i}$  the set obtained from $S$ by replacing the $i$-th sample with a new pair of head and tail entities drawn in the knowledge graph. Namely, $S^{i}=\{S\backslash\left(h_i,r_i,t_i\right)\cup \left(h_i',r_i,t_i'\right)\}$, where $\left(h_i',r_i,t_i'\right)$ is obtained by substituting $\left(h_i',t_i'\right)$ for $\left(h_i,t_i\right)$.

Since $\mathcal{A}$ can be regarded as a learning algorithm, we define $R(\mathcal{A},S)$ as its true risk or generalization error, which is a random variable on the training set $S$ and defined as
\[
\mathcal{R}(\mathcal{A},S)=
\mathds{E}_{z}[\mathcal{L}(\mathcal{A},z)].
\]
Here $\mathds{E}_{z}[\cdot]$ denotes the expectation when $z=\left(h,r,t\right)$ is a triple in knowledge graph and
$\mathcal{L}(\mathcal{A},z)$ is the loss function of the learning algorithm $\mathcal{A}$ with respect $z$.
It is well known that the true risk $\mathcal{R}(\mathcal{A},S)$ cannot be computed directly and is usually estimated by the empirical risk defined as
\[
\mathcal{R}_{emp}(\mathcal{A},S)=\frac{1}{n}\sum\limits_{k=1}^{n}\mathcal{L}(\mathcal{A},z_k),
\]
where $z_k=\left(h_k,r_k,t_k\right)$ is the $k$-th element of $S$.

After defining the two risks, the performance of learning algorithm $\mathcal{A}$ can be evaluated by the
difference between $\mathcal{R}(\mathcal{A},S)$ and $\mathcal{R}_{emp}(\mathcal{A},S)$, and we need to find the relation between the difference and the margin.
To this end, we define the Uniform-Replace-One stability motivated by \cite{bousquet2002stability}.
\begin{defi}
The learning algorithm $\mathcal{A}$ has Uniform-Replace-One stability $\beta$ with respect to the loss function $\mathcal{L}$
if for all $S\in \mathcal{Z}^n$ and  $i\in \{1,2,\ldots,n\}$, the following inequality holds
\[
\| \mathcal{L}(\mathcal{A}_S,\cdot)-\mathcal{L}(\mathcal{A}_{S^i},\cdot) \|_{\infty}\leq \beta.
\]
\end{defi}
Here $\mathcal{A}_S$ means that the learning algorithm $\mathcal{A}$ is trained on the data set $S$, $\|\mathcal{L}(\mathcal{A}_S,\cdot)\|_{\infty}$ is the maximum norm, and is equal to $\max_{z_k}\{\mathcal{L}(\mathcal{A}_S,z_k)\}$ for $k=1,2,\ldots,n$.
The loss function $\mathcal{L}$ of existing embedding methods takes the form
\begin{equation}
\mathcal{L}(\mathcal{A}_S,z)=f_r(h,t)+M-f_{r}(h',t'),\label{commonloss}
\end{equation}
where $\left(h,r,t\right), \left(h',r,t'\right) \in S$, $f_r(h,t)$ is a nonnegative score function. $M$ is the margin separating positive and negative
triples
and is usually set to be a global constant in a knowledge graph, e.g. $M=1$ in FB15K \cite{bordes2011learning}.
Moreover, $\mathcal{L}(\mathcal{A}_S,z)$ is defined as a nonnegative function, namely, if $f_r(h,t)+M-f_{r}(h',t')<0$, then we set $\mathcal{L}(\mathcal{A}_S,z)=0$.
 According to the loss function of embedding method $\mathcal{A}$, we have the following lemma.
\begin{lemma}\label{betabound}
The Uniform-Replace-One stability $\beta$ of the embedding methods with respect to the given loss function $\mathcal{L}(\mathcal{A}_S,z)$ is equal to $2\hat{f}_r$, where $\hat{f}_r=\max_{h,t} f_r(h,t)$ is the maximum over the triples $(h,r,t)\in S$.
\end{lemma}
\pf \ By the definition of $\beta$ and the loss function defined in Equation (\ref{commonloss}), we deduce
\begin{align*}
&\|\mathcal{L}(\mathcal{A}_S,\cdot)-\mathcal{L}(\mathcal{A}_{S^i},\cdot)\|_{\infty}\\
=&|f_r(h,t)+M-f_r(h',t')-f_r(h,t)-M+f_r(h'',t'')|\\
=&|f_r(h'',t'')-f_r(h',t')|\leq |f_r(h'',t'')|+|f_r(h',t')|\\
\leq& 2\max_{h,t} f_r(h,t).
\end{align*}
Setting $\hat{f}_r=\max_{h,t} f_r(h,t)$  completes the proof.
\qed

Now it is ready to reveal the relationship between the margin and the difference between the two risks of performance.
\begin{theo}\label{boundgeneral}
For the embedding method $\mathcal{A}$ with Uniform-Replace-One stability $\beta$ with respect to the given loss function $\mathcal{L}$, we have the following inequality with probability at least $1-\delta$,
\begin{equation}
\mathcal{R}(\mathcal{A},S)\leq \mathcal{R}_{emp}(\mathcal{A},S)+\sqrt{\frac{(M+\hat{f}_r)^2}{2n\delta}+\frac{6\hat{f}_r(M+\hat{f}_r)}{\delta}},
\end{equation}
where $\hat{f}_r=\max\limits_{h,t} f_r(h,t)$ is defined in Lemma \ref{betabound}.
\end{theo}

Before proving Theorem \ref{boundgeneral}, we first present the following Lemma verified in \cite{bousquet2002stability}.
\begin{lemma}\label{variancebound}
For any algorithm $\mathcal{A}$ and loss function $\mathcal{L}(\mathcal{A}_S,z)$ such that $0\leq \mathcal{L}(\mathcal{A}_S,z)\leq \hat{L}$, set $z_i=(h_i,r_i,t_i)\in S$, we have for any different $i,j\in \{1,2,\ldots,n\}$ that
\begin{align*}
&\mathds{E}_{S}\left[\left(\mathcal{R}(\mathcal{A},S)-\mathcal{R}_{emp}(\mathcal{A},S)\right)^2\right]\\
&\leq \frac{\hat{L}^2}{2n}+3\hat{L}\mathds{E}_{S\cup z_i'}[|\mathcal{L}(\mathcal{A}_S,z_i)-\mathcal{L}(\mathcal{A}_{S^i},z_i)|]
\end{align*}
\end{lemma}

\noindent \textbf{Proof of Theorem \ref{boundgeneral}:}
First, for the given loss function $\mathcal{L}(\mathcal{A}_S,z)$, we deduce that
\[
\mathcal{L}(\mathcal{A}_S,z)\leq M+f_r(h,t)\leq M+\hat{f}_r.
\]
Then from the definition of Uniform-Replace-One stability, we find that
$
\mathds{E}_{S\cup z_i'}[|\mathcal{L}(\mathcal{A}_S,z_i)-\mathcal{L}(\mathcal{A}_{S^i},z_i)|]\leq \beta.
$
Hence, by Lemma \ref{variancebound} and Lemma \ref{betabound}, we obtain that
\begin{align*}
&\mathds{E}_{S}\left[\left(\mathcal{R}(\mathcal{A},S)-\mathcal{R}_{emp}(\mathcal{A},S)\right)^2\right]\\
&\leq
\frac{(M+\hat{f}_r)^2}{2n}+6(M+\hat{f}_r)\hat{f}_r
\end{align*}
By Chebyshev's inequality, we derive that
\begin{align*}
&Prob(\left(\mathcal{R}(\mathcal{A},S)-\mathcal{R}_{emp}(\mathcal{A},S)\right)\geq \epsilon)\\
&\leq \frac{\mathds{E}_{S}\left[\left(\mathcal{R}(\mathcal{A},S)-\mathcal{R}_{emp}(\mathcal{A},S)\right)^2\right]}{\epsilon^2}\\
&\leq
\left(\frac{(M+\hat{f}_r)^2}{2n}+6(M+\hat{f}_r)\hat{f}_r\right).
\end{align*}
Let the right hand side of the above inequality be $\delta$, then we have with probability at least $1-\delta$
that
\[
\mathcal{R}(\mathcal{A},S)\leq \mathcal{R}_{emp}(\mathcal{A},S)+\sqrt{\frac{(M+\hat{f}_r)^2}{2n\delta}+\frac{6\hat{f}_r(M+\hat{f}_r)}{\delta}}.
\]
This completes the proof.
\qed

From Theorem \ref{boundgeneral}, it can be seen that a large margin would lead to over-fitting during the learning process. This is reasonable since a large margin means more incorrect triples $(h',r,t')$ involve in the computation of the loss function (\ref{commonloss}) such that $f_r(h,t)+M-f_r(h',t')\geq 0$.
However, setting $M$ as small as possible is also inadvisable since it is a strict constraint that excludes incorrect triples whose values $f_r(h',t')$ are slightly larger than $f_r(h,t)$. Therefore, it demands a strategy to choose moderately small values of $M$ in practice so as to make the error of performance as small as possible.

\section{Locally Adaptive Translation Method}

In the previous section, to obtain better performance of embedding, it has proved that it is necessary to find an appropriate loss function in terms of margin over different knowledge graphs.
In this section, we propose a locally adaptive translation method, called TransA, to adaptively choose the optimal margin.

Because the classical knowledge graph is fully made up of two disjoint sets, i.e., the entity set and relation set, it makes sense that
the optimal margin, denoted by $M_{opt}$, is composed of two parts, namely, entity-specific margin $M_{ent}$, and relation-specific margin $M_{rel}$. Furthermore, it is natural to linearly combine the two specific margins via a parameter $\mu$ which controls the trade-off between them.
Therefore, the optimal margin of embedding satisfies
\begin{equation}
M_{opt}=\mu M_{ent}+(1-\mu)M_{rel},\label{totalmargin}
\end{equation}
where $0\leq \mu\leq 1$.
To demonstrate that the margin $M_{opt}$ is optimal, it is sufficient to find the optimal entity-specific margin and the optimal relation-specific margin, and we will elaborate this in the following.

\subsection{Entity-specific margin}

\begin{figure}[h,t]
\centering
\setlength{\abovecaptionskip}{4pt}
\setlength{\belowcaptionskip}{-10pt}
\scalebox{0.5}{\includegraphics{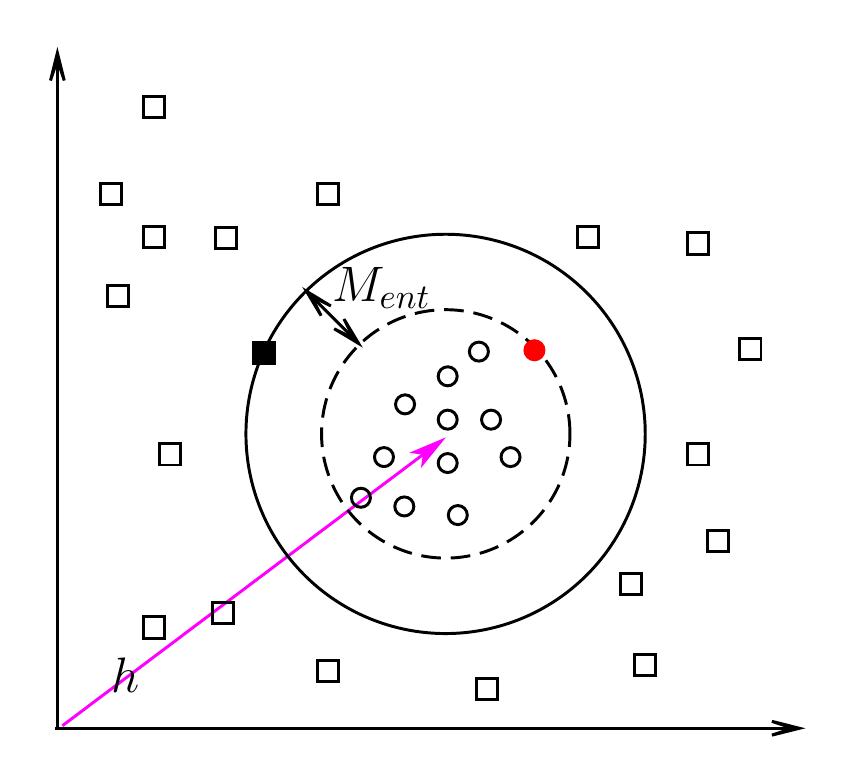}}
\caption{The illustration of the entity-specific margin $M_{ent}$. The points marked by circle and rectangle are positive and negative entities, respectively.}
\label{marill}
\end{figure}

To define the optimal entity-specific margin $M_{ent}$, it  has been verified by \cite{fan2015large} that for a specific head entity $h$ (or tail entity $t$), the best performance is achieved when the  embedding of entities brings the positive tail entities (or head entities)  close to each other, and moves the negative ones with a margin. The positive entities have the same relation with $h$ (or $t$), and the negative entities have different relations with $h$ (or $t$).
In this sense, the optimal margin $M_{ent}$ is actually equal to the distance between two concentric spheres in the vector space, illustrated in Figure \ref{marill}.
The positive entities (illustrated as ``$\bigcirc$'') are constrained within the internal sphere,
while the negative entities (illustrated as ``$\square$") lie outside the external sphere.
The interpretation of $M_{ent}$ is motivated by the work of metric learning, such as \cite{weinberger2009distance,do2012metric}.
Notice that the above analysis applies to both the head entity $h$ and the tail entity $t$, thus we simply use entity $h$ to stand for both cases in the rest of the paper.

More formally, for a specific entity $h$ and one related relation $r$, the sets of positive and negative entities with respect to $r$ are defined as $P_r=\{t|(h,r,t)\in \Delta\}$ and $N_r=\{t|(h,r,t)\not \in \Delta, (h,r',t)\in \Delta, \exists r'\in R\}$, where $\Delta$ is the set of correct triples,
$R$ is the set of relations in the knowledge graph. In other words, the set of negative entities $N_r$ contains those which have relations with $h$ of type other than $r$.
Remark that due to the multi-relational property of the knowledge graph, it may be true that $P_r\subseteq N_r$.
Then the optimal margin $M_{ent}$ is equal to the distance between the external sphere and the internal sphere, which push the elements in $N_r$ away from  $h$ and keep the elements in $P_r$ close to each other.
Let $nr_h$ be the number of relations of the entity $h$, i.e., the number of different types of relations with $h$ as one end.
Set $R_h$ be the set of relations related to $h$. More formally, we define $M_{ent}$ as follows.
\begin{defi}[Entity-specific margin]\label{emar}
For a given entity $h$, for all $t\in P_r$ and $t'\in N_r$,
\[
M_{ent}=\frac{\sum\limits_{r\in R_h} \min\limits_{t,t'}\sigma(\|h-t'\|-\|h-t\|)}{nr_h},
\]
where
\[
\sigma(x)=\left\{
    \begin{array}{ll}
    x &\mbox{when $x\geq 0$;}\\
    -x &\mbox{otherwise.}
    \end{array}
    \right.
\]
returns the absolute value of $x$.
\end{defi}

It can be seen from Definition \ref{emar} that for each relation $r$, the value $\min_{t,t'}\sigma(\|h-t'\|-\|h-t\|)$ obtains the minimum when it takes the nearest negative entity and the farthest positive entity with respect to $h$. In Figure \ref{marill}, the nearest negative entity is marked as the black rectangle, and the farthest positive entity is marked as the red circle.
This definition is kind of similar to the margin defined in Support Vector Machine~\cite{vapnik2013nature,boser1992training}, where the margin of two classes with respect to the classification hyperplane is equal to the minimum absolute difference of the distances of any two different-class instances projected to the norm vector of the hyperplane.
In particular, when $N_r=\emptyset$, we set
$M_{ent}=0$, which is reasonable since all positive entities are within the internal sphere.
Remark that the reason we consider the term $nr_h$ as a denominator is to discriminate relations with different mapping properties, i.e., 1-to-1, 1-to-N, N-to-1, N-to-N.
Besides, in order to apply the optimal margin $M_{ent}$ to large data set, we similarly adopt the active set method proposed in \cite{weinberger2008fast} to speed up the calculation.
We check a very small fraction of negative entities typically lying nearby the farthest positive entity for several rounds. Averaging the margins obtained in each round leads to the final entity-specific margin.

\subsection{Relation-specific margin}

\begin{figure}[h,t]
\centering
\setlength{\abovecaptionskip}{4pt}
\setlength{\belowcaptionskip}{-18pt}
\scalebox{0.5}{\includegraphics{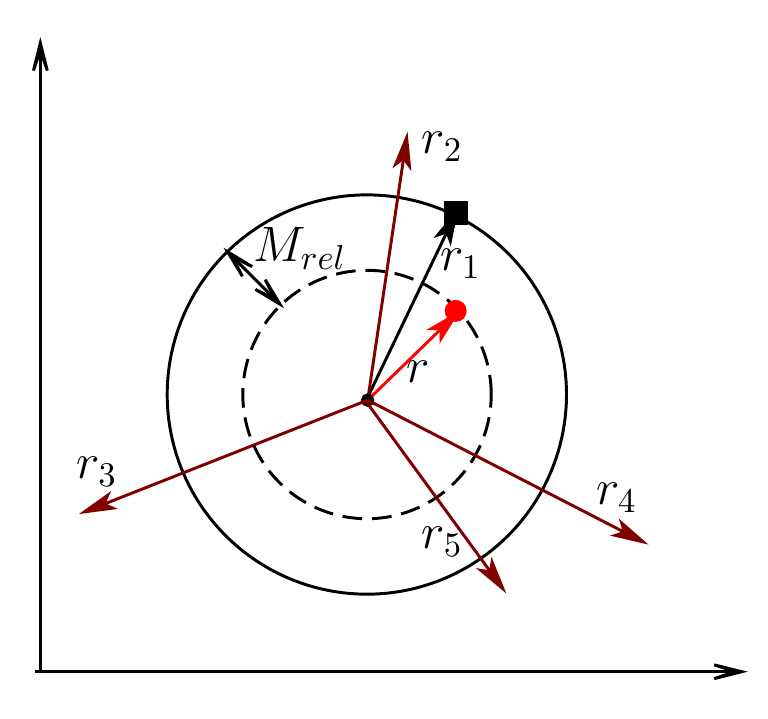}}
\caption{The illustration of the relation-specific margin $M_{rel}$.}
\label{marillrel}
\end{figure}

The optimal relation-specific margin $M_{rel}$ from the relation aspect is found by considering the proximity of relations concerning a given entity. For a specific entity $h$ and one of its related relation $r$, if $N_r\neq \emptyset$, set $R_{h,r}=\{r_1,r_2,\ldots,r_{nh_{r}-1}\}$ be the set of relations the entity $h$ has except $r$.
Different relations in $R_{h,r}$ have different degrees of similarity with the relation $r$.
To measure this similarity, we consider the length of relation-specific embedding vectors.
We classify the relations in $R_{h,r}$ into two parts according to whether its length is larger than $\|r\|$.
For relations $r_i,r_j\in R_{h,r}$ with lengths larger than  $\|r\|$, we assume that $r_i$ is more similar with $r$ than $r_j$ if $\|r_i\|-\|r\|\leq \|r_j\|-\|r\|$. Then similar to the analysis of entity-specific margin, the optimal relation-specific margin is equal to the distance between two concentric spheres in the vector space. The internal sphere constraints relation $r$ and those with length smaller than $\|r\|$, while the relations with length greater than $\|r\|$ lie outside the external sphere.
More formally, we define the optimal relation-specific margin as follows.
\begin{defi}[Relation-specific margin]
For a given entity $h$ and one of its related relations $r$, if \ $\exists\  r_i\in R_{h,r}$ such that $\|r_i\|\geq \|r\|$, then
\begin{equation}
M_{rel}=\min\limits_{r_i \in R_{h,r}}(\|r_i\|-\|r\|).
\end{equation}
\end{defi}
It follows from the definition of $M_{rel}$ that  for $r_i\in R_{h,r}$ with $\|r_i\|\geq \|r\|$, we have
$\|r_i\|-\|r\|\geq M_{rel}$. In other words, $\|r_i\|\geq M_{rel}+\|r\|$ holds for those $r_i\in R_{h,r}$.
This means that $M_{rel}$ is determined by the most similar relation(s) with smallest difference of length,  and it pushes other dissimilar relations whose lengths are much larger than $\|r\|$ away from $r$. In this sense, $M_{rel}$ is optimal. In Figure \ref{marillrel}, we illustrate the calculation of $M_{rel}$ with $M_{rel}=\|r_1\|-\|r\|$. It can be seen that any of the other relations $r_i$ ($i=2,3,4,5$) has length larger than $r_1$.
In particular, if $R_{h,r}=\emptyset$ or no $r_i\in R_{h,r}$ exists such that $\|r_i\|\geq \|r\|$, then we set $M_{rel}=0$ by convention. This makes sense since relations with length smaller than $\|r\|$ are inherently constrained within the internal sphere of radius $\|r\|$.

\subsection{The locally adaptive method TransA}

Given a triple $(h,r,t)$, we define the margin-varying objective function as follows.
\[
\sum\limits_{(h,r,t)\in \Delta}\sum\limits_{(h',r,t')\in \Delta'}\max(0,f_r(h,t)+M_{opt}-f_{r}(h',t')),
\]
where $\max(x,y)$ returns the maximum between $x$ and $y$, $M_{opt}=\mu M_{ent}+(1-\mu)M_{rel}$ is the optimal margin defined by Equation (\ref{totalmargin}), $\Delta$ is the set of correct triples and $\Delta'$ is the set of incorrect triples.
Following the construction of incorrect triples \cite{bordes2013translating}, we replace the head or tail entities in the correct triple $(h,r,t)\in \Delta$ to obtain
$(h',r,t')\in \Delta'$.

In fact, the optimal margin defined in Equation (\ref{totalmargin}) from the entity and relation aspects provides a way to characterize the locality of knowledge graph in that it considers the structure information in the graph. For one thing, the entity-specific margin models the local distance between negative and positive entities. For another, the relation-specific margin quantifies the proximity of relations. Moreover, when the entities and relations change, the optimal margin varies accordingly, without fixing its value among some predefined candidates.
This is way the method is called locally adaptive.

The learning process of TransA employs the widely used stochastic gradient descent (SGD) method. To combat over-fitting, we follow the initialization of entity and relation embeddings as in \cite{bordes2013translating}.
In this paper, without loss of generality, we assume that given a triple $(h,r,t)$, the score function $f_r(h,t)$ takes the form as
\begin{equation}
f_r(h,t)=\|h+r-t\|,\label{standscore}
\end{equation}
where $\|\cdot\|$ represents the $L_1$-norm or $L_2$-norm of the vector $h+r-t$, and $h,r,t\in \mathbb{R}^d$, $d$ is the dimension of the embedded vector space. Remark that although we suppose that entity and relation embedding are  in the same vector space $\mathbb{R}^d$, it is not difficult to extend the method to that in different vector spaces similar to the work \cite{lin2015learning}.

\section{Experiments}

We will conduct experiments on two tasks: link prediction \cite{bordes2013translating} and triple classification \cite{wang2014knowledge}.
The data sets we use are publicly available from two widely used knowledge graphs, WordNet \cite{miller1995wordnet} and Freebase \cite{bollacker2008freebase}. For the data sets from WordNet, we employ WN18 used in \cite{bordes2014semantic} and WN11 used in \cite{socher2013reasoning}. For the data sets of Freebase, we employ FB15K used also in \cite{bordes2014semantic} and FB13 used in \cite{socher2013reasoning}. The statistics of these data sets are listed in Table \ref{direct}.

\begin{table}[h,t]
\centering
\setlength{\abovecaptionskip}{4pt}
\setlength{\belowcaptionskip}{-10pt}
\small
\begin{tabular}{|c|c|c|c|c|c|}
\hline
Data sets & \#Rel &  \#Ent & \#Train & \#Valid & \#Test\\
\hline
WN18 &  18& 40,943 & 141,442 & 5,000 & 5,000 \\
FB15K &  1,345& 14,951 & 483,142 & 50,000 & 59,071 \\
WN11 &  11& 38,696 & 112,581 & 2,609 & 10,544\\
FB13 & 13& 75,043 & 316,232 & 5,908 & 23,733 \\
\hline
\end{tabular}
\caption{The data sets.}
\label{direct}
\end{table}

\subsection{Link prediction}

Link prediction aims to predict the missing entities $h$ or $t$ for a triple $(h,r,t)$. Namely, it predicts $t$ given $(h,r)$ or predict $h$ given $(r,t)$. Similar to the setting in \cite{bordes2011learning,bordes2013translating,zhao2014content}, the task returns a list of candidate entities from the knowledge graph instead of one best answer, and we conduct the link prediction task on the two data sets WN18 and FB15K.
Following the procedure used in \cite{bordes2013translating}, we also adopt
the evaluation measure, namely, mean rank (i.e., mean rank of correct entities).
It is clear that a good predictor has lower mean rank.
In the test stage, for each test triple $(h,r,t)$, we replace the head or tail entity by all entities in the knowledge graph, and rank the entities in the decreasing order with respect to the scores calculated by $f_r$.
To distinguish the corrupted triples which are also correct ones, we also filter out the corrupted triples before
the ranking of candidate entities. This operation is denoted as ``filter'' and ``raw'' otherwise.

The baseline methods include classical embedding methods, such as TransE~\cite{bordes2013translating}, TransH~\cite{wang2014knowledge}, and others shown in Table \ref{directmr}.
Since the data sets we used are the same as our baselines, we compare our results with them reported in \cite{wang2014knowledge}. The  learning rate $\lambda$ during the SGD process is selected among $\{0.1,0.01,0.001\}$, the embedding dimension $d$ in $\{20,50,100\}$, the batch size $B$ among $\{20,120,480,1440,4800\}$, and the parameter $\mu$ in Equation (\ref{totalmargin}) in $[0,1]$.
Notice that the optimal margin for TransA is not predefined but computed by Equation (\ref{totalmargin}).
All parameters are determined on the validation set.
The optimal settings are: $\lambda=0.001$, $d=100$, $B=1440$, $\mu=0.5$ and taking $L_1$ as dissimilarity on WN18;
$\lambda=0.001$, $d=50$, $B=4800$, $\mu=0.5$ and taking $L_1$ as dissimilarity on FB15K.

Experiment results are shown in Table \ref{directmr}. It can be seen that on both data sets, TransA obtains the lowest mean rank. Furthermore, on WN18, among the baselines, Unstructured and TransH(unif) perform the best, but TransA decreases the mean rank by about 150 compared with both of them.
On FB15K, among the baselines, TransH(unif) is the best baseline. TransA decreases its mean rank by $30\sim 50$.
Notice that the decreases on WN18 and FB15K are different, because the number of relations in WN18 is quite small and the relation-specific margin is very small too. In this case, the optimal margin is almost equal to the entity-specific margin. While on FB15K, the number of relations is 1345, and the optimal margin is the combination of the entity-specific margin and the relation-specific margin.

\begin{table}[h,t]
\centering
\setlength{\abovecaptionskip}{4pt}
\setlength{\belowcaptionskip}{-18pt}
\begin{tabular}{|c|c|c|c|c|}
\hline
Data sets & \multicolumn{2}{c}{WN18} &  \multicolumn{2}{|c|}{FB15K}\\
\hline
\multirow{2}{*}{Metric} & \multicolumn{2}{|c}{Mean Rank} &  \multicolumn{2}{|c|}{Mean Rank}\\
\cline{2-5}
 & Raw & Filter & Raw & Filter \\
\hline
\cline{1-5}
Unstructured & 315  &304&	1,074&	979\\
RESCAL & 1,180  &1,163&	828&	683\\
SE & 1,011  &985&	273&	162\\
SME(linear) & 545  &533&	274&	154\\
SME(bilinear) & 526  &509&	284&	158\\
LFM & 469  &456&	283&	164\\
TransE &263&251& 243 &125
  \\
TransH(bern)&401 &388 &212 &87\\
TransH(unif)& 318 &303 &211 &84 \\
\hline
TransA & 165  & 153 & 164	 &58	\\
\hline
\end{tabular}
\caption{Evaluation results on link prediction.}\label{directmr}
\end{table}

\subsection{Triple classification}

Triple classification, studied in \cite{socher2013reasoning,wang2014knowledge}, is to confirm whether a triple $(h,r,t)$ is correct or not, namely, a binary classification problem on the triple.
The data sets we use in this task are WN11 and FB13 used in \cite{socher2013reasoning} and FB15K used in \cite{wang2014knowledge}.
Following the evaluation in NTN \cite{socher2013reasoning}, the evaluation needs negative labels. The data sets WN11 and FB13 already have negative triples, which are obtained by corrupting golden ones. For FB15K, we follow the same way to construct negative triples as \cite{socher2013reasoning}. The classification is evaluated as follows. For a triple $(h,r,t)$, if the dissimilarity score obtained by $f_r$ is less than a relation-specific threshold, then the triple is classified to be positive, and negative otherwise. The threshold is determined by maximizing accuracy on the validation data set.

The baseline methods include classical embedding methods, such as TransE~\cite{bordes2013translating}, TransH~\cite{wang2014knowledge}, and others shown in Table \ref{comptc}.
The learning rate $\lambda$ during the stochastic gradient descent process is selected among $\{0.1,0.01,0.001\}$, the embedding dimension $d$ in $\{20,50,100,200,220,300\}$, the parameter $\mu$ in Equation (\ref{totalmargin}) in $[0,1]$, and batch size $B$ among $\{20,120,480,1440,4800\}$.
The optimal margin for TransA is not predefined but computed according to Equation (\ref{totalmargin}).
All parameters are determined on the validation set.
The optimal setting are: $\lambda=0.001$, $d=220$, $B=120$, $\mu=0.5$ and taking $L_1$ as dissimilarity on WN11;
$\lambda=0.001$, $d=50$, $B=480$, $\mu=0.5$ and taking $L_1$ as dissimilarity on FB13.

Experiment results are shown in Table \ref{comptc}.
On WN11, TransA outperforms the other methods. On FB13, the method NTN is shown more powerful. This is consistent with the results in previous literature \cite{wang2014knowledge}. On FB15K, TransA also performs the best.
Since FB13 is much denser to FB15K, NTN is more expressive on dense graph. On sparse graph, TransA is superior to other state-of-the-art embedding methods.

\begin{table}[h,t]
\centering
\setlength{\abovecaptionskip}{4pt}
\setlength{\belowcaptionskip}{-18pt}
\begin{tabular}{|c|c|c|c|}
\hline
Data sets & WN11 &  FB13 & FB15K\\
\hline
SE & 53.0 & 75.2 & -\\
SME(linear) & 70.0 &63.7 & -\\
SLM &69.9 &85.3 &-\\
LFM &73.8 &84.3 &-\\
NTN &70.4 &87.1 &68.5\\
TransH(unif)& 77.7 &76.5 &79.0\\
TransH(bern)& 78.8 &83.3 &80.2\\
\hline
TransA & 93.2  & 82.8 &	87.7 \\
\hline
\end{tabular}
\caption{Evaluation results of triple classification. (\%)}\label{comptc}
\end{table}

\section{Conclusion}

In this paper, we tackled the knowledge embedding problem and proposed a locally adaptive translation method, called TransA, to adaptively learn the representation of entities and relations in a knowledge graph. We firstly presented the necessity of choosing an appropriate margin in the margin-based loss function. Then we defined the optimal margin from the entity and relation aspects, and integrated the margin into the commonly used loss function for knowledge embedding. Experimental results validate the effectiveness of the proposed method.

\section{Acknowledgement}

This work is supported by National Grand Fundamental Research 973 Program of China (No. 2012CB316303, 2014CB340401), National Natural Science Foundation of China (No. 61402442, 61572469, 61572473, 61303244), Beijing Nova Program (No. Z121101002512063), and Beijing Natural Science Foundation (No. 4154086).

\bibliographystyle{aaai}
\bibliography{jia}

\begin{thebibliography}{}

\bibitem[\protect\citeauthoryear{Bollacker \bgroup et al\mbox.\egroup
  }{2008}]{bollacker2008freebase}
Bollacker, K.; Evans, C.; Paritosh, P.; Sturge, T.; and Taylor, J.
\newblock 2008.
\newblock Freebase: a collaboratively created graph database for structuring
  human knowledge.
\newblock In {\em Proceedings of the 2008 ACM SIGMOD international conference
  on Management of data},  1247--1250.
\newblock ACM.

\bibitem[\protect\citeauthoryear{Bordes \bgroup et al\mbox.\egroup
  }{2011}]{bordes2011learning}
Bordes, A.; Weston, J.; Collobert, R.; and Bengio, Y.
\newblock 2011.
\newblock Learning structured embeddings of knowledge bases.
\newblock In {\em Conference on Artificial Intelligence}, number
  EPFL-CONF-192344.

\bibitem[\protect\citeauthoryear{Bordes \bgroup et al\mbox.\egroup
  }{2012}]{bordes2012joint}
Bordes, A.; Glorot, X.; Weston, J.; and Bengio, Y.
\newblock 2012.
\newblock Joint learning of words and meaning representations for open-text
  semantic parsing.
\newblock In {\em International Conference on Artificial Intelligence and
  Statistics},  127--135.

\bibitem[\protect\citeauthoryear{Bordes \bgroup et al\mbox.\egroup
  }{2013}]{bordes2013translating}
Bordes, A.; Usunier, N.; Garcia-Duran, A.; Weston, J.; and Yakhnenko, O.
\newblock 2013.
\newblock Translating embeddings for modeling multi-relational data.
\newblock In {\em Advances in Neural Information Processing Systems},
  2787--2795.

\bibitem[\protect\citeauthoryear{Bordes \bgroup et al\mbox.\egroup
  }{2014}]{bordes2014semantic}
Bordes, A.; Glorot, X.; Weston, J.; and Bengio, Y.
\newblock 2014.
\newblock A semantic matching energy function for learning with
  multi-relational data.
\newblock {\em Machine Learning} 94(2):233--259.

\bibitem[\protect\citeauthoryear{Boser, Guyon, and
  Vapnik}{1992}]{boser1992training}
Boser, B.~E.; Guyon, I.~M.; and Vapnik, V.~N.
\newblock 1992.
\newblock A training algorithm for optimal margin classifiers.
\newblock In {\em Proceedings of the fifth annual workshop on Computational
  learning theory},  144--152.
\newblock ACM.

\bibitem[\protect\citeauthoryear{Bousquet and
  Elisseeff}{2002}]{bousquet2002stability}
Bousquet, O., and Elisseeff, A.
\newblock 2002.
\newblock Stability and generalization.
\newblock {\em The Journal of Machine Learning Research} 2:499--526.

\bibitem[\protect\citeauthoryear{Do \bgroup et al\mbox.\egroup
  }{2012}]{do2012metric}
Do, H.; Kalousis, A.; Wang, J.; and Woznica, A.
\newblock 2012.
\newblock A metric learning perspective of svm: on the relation of lmnn and
  svm.
\newblock In {\em International Conference on Artificial Intelligence and
  Statistics},  308--317.

\bibitem[\protect\citeauthoryear{Fan \bgroup et al\mbox.\egroup
  }{2015}]{fan2015large}
Fan, M.; Zhou, Q.; Zheng, T.~F.; and Grishman, R.
\newblock 2015.
\newblock Large margin nearest neighbor embedding for knowledge representation.
\newblock {\em arXiv preprint arXiv:1504.01684}.

\bibitem[\protect\citeauthoryear{Jenatton \bgroup et al\mbox.\egroup
  }{2012}]{jenatton2012latent}
Jenatton, R.; Roux, N.~L.; Bordes, A.; and Obozinski, G.~R.
\newblock 2012.
\newblock A latent factor model for highly multi-relational data.
\newblock In {\em Advances in Neural Information Processing Systems},
  3167--3175.

\bibitem[\protect\citeauthoryear{Jia \bgroup et al\mbox.\egroup
  }{2014}]{jia2014openkn}
Jia, Y.; Wang, Y.; Cheng, X.; Jin, X.; and Guo, J.
\newblock 2014.
\newblock Openkn: An open knowledge computational engine for network big data.
\newblock In {\em Advances in Social Networks Analysis and Mining (ASONAM),
  2014 IEEE/ACM International Conference on},  657--664.
\newblock IEEE.

\bibitem[\protect\citeauthoryear{Lin \bgroup et al\mbox.\egroup
  }{2015}]{lin2015learning}
Lin, Y.; Liu, Z.; Sun, M.; Liu, Y.; and Zhu, X.
\newblock 2015.
\newblock Learning entity and relation embeddings for knowledge graph
  completion.
\newblock In {\em Proceedings of AAAI}.

\bibitem[\protect\citeauthoryear{Liu \bgroup et al\mbox.\egroup
  }{2014}]{liu2014lsdh}
Liu, D.; Wang, Y.; Jia, Y.; Li, J.; and Yu, Z.
\newblock 2014.
\newblock Lsdh: a hashing approach for large-scale link prediction in
  microblogs.
\newblock In {\em Twenty-Eighth AAAI Conference on Artificial Intelligence}.

\bibitem[\protect\citeauthoryear{Miller}{1995}]{miller1995wordnet}
Miller, G.~A.
\newblock 1995.
\newblock Wordnet: a lexical database for english.
\newblock {\em Communications of the ACM} 38(11):39--41.

\bibitem[\protect\citeauthoryear{Nickel, Tresp, and
  Kriegel}{2012}]{nickel2012factorizing}
Nickel, M.; Tresp, V.; and Kriegel, H.-P.
\newblock 2012.
\newblock Factorizing yago: scalable machine learning for linked data.
\newblock In {\em Proceedings of the 21st international conference on World
  Wide Web},  271--280.
\newblock ACM.

\bibitem[\protect\citeauthoryear{Socher \bgroup et al\mbox.\egroup
  }{2013}]{socher2013reasoning}
Socher, R.; Chen, D.; Manning, C.~D.; and Ng, A.
\newblock 2013.
\newblock Reasoning with neural tensor networks for knowledge base completion.
\newblock In {\em Advances in Neural Information Processing Systems},
  926--934.

\bibitem[\protect\citeauthoryear{Sutskever, Tenenbaum, and
  Salakhutdinov}{2009}]{sutskever2009modelling}
Sutskever, I.; Tenenbaum, J.~B.; and Salakhutdinov, R.~R.
\newblock 2009.
\newblock Modelling relational data using bayesian clustered tensor
  factorization.
\newblock In {\em Advances in neural information processing systems},
  1821--1828.

\bibitem[\protect\citeauthoryear{Vapnik}{2013}]{vapnik2013nature}
Vapnik, V.
\newblock 2013.
\newblock {\em The nature of statistical learning theory}.
\newblock Springer Science \& Business Media.

\bibitem[\protect\citeauthoryear{Wang \bgroup et al\mbox.\egroup
  }{2014}]{wang2014knowledge}
Wang, Z.; Zhang, J.; Feng, J.; and Chen, Z.
\newblock 2014.
\newblock Knowledge graph embedding by translating on hyperplanes.
\newblock In {\em Proceedings of the Twenty-Eighth AAAI Conference on
  Artificial Intelligence},  1112--1119.
\newblock Citeseer.

\bibitem[\protect\citeauthoryear{Weinberger and
  Saul}{2008}]{weinberger2008fast}
Weinberger, K.~Q., and Saul, L.~K.
\newblock 2008.
\newblock Fast solvers and efficient implementations for distance metric
  learning.
\newblock In {\em Proceedings of the 25th international conference on Machine
  learning},  1160--1167.
\newblock ACM.

\bibitem[\protect\citeauthoryear{Weinberger and
  Saul}{2009}]{weinberger2009distance}
Weinberger, K.~Q., and Saul, L.~K.
\newblock 2009.
\newblock Distance metric learning for large margin nearest neighbor
  classification.
\newblock {\em The Journal of Machine Learning Research} 10:207--244.

\bibitem[\protect\citeauthoryear{Wu \bgroup et al\mbox.\egroup
  }{2012}]{wu2012probase}
Wu, W.; Li, H.; Wang, H.; and Zhu, K.~Q.
\newblock 2012.
\newblock Probase: A probabilistic taxonomy for text understanding.
\newblock In {\em Proceedings of the 2012 ACM SIGMOD International Conference
  on Management of Data},  481--492.
\newblock ACM.

\bibitem[\protect\citeauthoryear{Zhao, Jia, and Wang}{2014}]{zhao2014content}
Zhao, Z.; Jia, Y.; and Wang, Y.
\newblock 2014.
\newblock Content-structural relation inference in knowledge base.
\newblock In {\em Twenty-Eighth AAAI Conference on Artificial Intelligence}.

\end{thebibliography}

\end{document}